\documentclass[11pt,a4paper]{article}
\usepackage[hyperref]{emnlp2021}
\usepackage{danudefs}
\usepackage{times}
\usepackage{latexsym}
\usepackage{graphicx}
\usepackage{amsmath, amsfonts, amssymb, amsthm, mathtools}
\usepackage{dsfont}
\usepackage{booktabs}
\usepackage{algorithm, algorithmic}

\usepackage{booktabs}
\usepackage{microtype}

\title{Detect and Classify -- Joint Span Detection and Classification \\for Health Outcomes}

\author{Micheal Abaho$^1$
  \\\And
  Danushka Bollegala$^{1,2}$\Thanks{ Danushka Bollegala holds concurrent appointments as a Professor at University of Liverpool and as an Amazon Scholar. This paper describes work performed at the University of Liverpool and is not associated with Amazon.}\\
    \kern+10em $^1$University of Liverpool, United Kingdom \\
     \kern+10em $^2$Amazon \\
  \kern+10em \texttt{\{m.abaho,danushka,prw,shinds\}@liverpool.ac.uk} 
  \\\And
  Paula Williamson$^1$ \\
  \\\And
  Susanna Dodd$^1$ \\
  \\}

\begin{document}
\maketitle
\begin{abstract}

A health outcome is a measurement or an observation used to capture and assess the effect of a treatment. Automatic detection of health outcomes from text would undoubtedly speed up access to evidence necessary in healthcare decision making. Prior work on outcome detection has modelled this task as either (a) a \emph{sequence labelling task}, where the goal is to detect which text spans describe health outcomes,
or (b) a \emph{classification task}, where the goal is to classify a text into a pre-defined set of categories depending on an outcome that is mentioned somewhere in that text. 
However, this decoupling of span detection and classification is problematic from a modelling perspective and ignores global structural correspondences between sentence-level and word-level information present in a given text.
To address this, we propose a method that uses both word-level and sentence-level information to \emph{simultaneously} perform outcome span detection and outcome type classification.
In addition to injecting contextual information to hidden vectors, we use label attention to appropriately weight both word and sentence level information. 
Experimental results on several benchmark datasets for health outcome detection show that our proposed method consistently outperforms decoupled methods, reporting competitive results.
\end{abstract}

\section{Introduction}

\begin{table}[t]
\resizebox{\columnwidth}{!}{
\begin{tabular}{@{}ll@{}}
\toprule
sentence                                                    & \begin{tabular}[c]{@{}l@{}}There were no significance between-\\ group differences in the incidence of \\ \textbf{\textcolor{Green}{wheezing}} or \textbf{\textcolor{Green}{shortness of breath}}\end{tabular}                                                                                                           \\ \midrule
OSD                                                         & \begin{tabular}[c]{@{}l@{}}Outcomes: \textbf{wheezing}, \\                   \textbf{shortness of Breath}\end{tabular}                                                                                                                                                             \\ \midrule
OC                                                          & Outcome type: \colorbox{Green}{Physiological}                                                                                                                                                                                                                                      \\ \midrule
\begin{tabular}[c]{@{}l@{}}Joint OSD \\ \& OC\end{tabular} & \begin{tabular}[c]{@{}l@{}}Outcomes-Outcome type\\ \textbf{wheezing}-\colorbox{Green}{Physiological}\\ \textbf{Shortness of Breath}-\colorbox{Green}{Physiological}\end{tabular}                                                                                                                                       \\ \midrule\midrule
sentence                                                    & \begin{tabular}[c]{@{}l@{}}Cumulative incidence and relative \\ risks with 95\% confidence intervals\\ for \textbf{\textcolor{BlueGreen}{death from any cause}}, \textbf{\textcolor{BlueGreen}{death from}} \\ \textbf{\textcolor{BlueGreen}{prostate cancer}}, and metastasis\\ were estimated in intention-to-treat \\ and per-protocol analyses.\end{tabular} \\ \midrule
OSD                                                         & \begin{tabular}[c]{@{}l@{}}Outcomes: \textbf{death from any cause},\\                   \textbf{death from prostate cancer}\end{tabular}                                                                                                                                           \\ \midrule 
OC                                                          & Outcome type: \colorbox{BlueGreen}{Mortality}                                                                                                                                                                                                                                          \\ \midrule
\begin{tabular}[c]{@{}l@{}}Joint OSD\\ \& OC\end{tabular}   & \begin{tabular}[c]{@{}l@{}}Outcomes-Outcome type\\ \textbf{death from any cause}-\colorbox{BlueGreen}{Mortality}\\ \textbf{death from prostate cancer}-\colorbox{BlueGreen}{Mortality} \end{tabular}                                                                                                                            \\ \bottomrule
\end{tabular}
}
\caption{Comparing the output of the three separate HOD tasks given two sample sentences. OSD retrieves the outcome spans, OC classifies
the text span into a set of outcome types, and Joint OSD \& OC retrieves outcomes and classifies them into outcome types.}
\label{tab:example}
\end{table}

Access to the best available evidence in context of patient's individual conditions enables healthcare professionals to administer optimal patient care~\cite{demner2006automatically}. 
Healthcare professionals identify outcomes as a fundamental part of the evidence they require to make decisions~\cite{van-aken-etal-2021-clinical}.
\newcite{Williamson2017The1.0} define an outcome as a measurement or an observation used to capture and assess the effect of treatment such as assessment of side effects (risk) or effectiveness (benefits). 
With the rapid growth of literature that reports outcomes, researchers have acknowledged and addressed the need to automate the extraction of outcomes from systematic reviews \cite{jonnalagadda2015automating, Nye2018ALiterature} and answering clinical questions \cite{demner2007answering}. 
\newcite{jin2018advancing} mention that automated Health Outcomes Detection (HOD) could speed up the process of analysing and assessing the effectiveness of clinical interventions in Evidence Based Medicine~\cite[EBM;][]{Sackett71}.

HOD has been conducted in the past as either an \emph{Outcome Span Detection} (OSD) task, where we must detect a continuous span of tokens indicating a health outcome ~\cite{Nye2018ALiterature, brockmeier2019improving} or as an \emph{Outcome Classification} (OC) task, where the goal is to classify the text spans into a pre-defined set of categories~\cite{wallace2016extracting, jin2018advancing, kiritchenko2010exact}. 
However, the two tasks are highly correlated and local token-level information enables us to make accurate global sentence-level outcome predictions, and vice versa.
An outcome type predicted for a text span in a sentence must be consistent with the other outcome spans detected from the same sentence, while the outcome spans detected from a sentence must be compatible with their outcome types.
These mutual compatibility constraints between outcome spans and their classes will be lost in a decoupled approach, resulting in poor performance for both OSD and OC tasks.

Two illustrative examples in \autoref{tab:example} show the distinction between the OSD, OC and Joint OSD \& OC tasks. Specifically, in the first sentence, OSD extracts all outcomes i.e. \textit{wheezing} and \textit{shortness of breath}, OC classifies the text into an outcome type, Physiological, and then Joint OSD \& OC extracts an outcome span and classifies it concurrently i.e. it extracts \textit{wheezing} and also classifies it as a Physiological outcome. 
Motivated by the recent success in joint modelling of tasks such as aspect extraction (AE) and aspect sentiment classification (ASC), which together make a customer sentiment analysis task called Aspect Based Sentiment Analysis~\cite[ABSA;][]{xu2019bert}, we model HOD as a joint task involving both OSD and OC.
HOD can be formally defined as follows:
\paragraph{Health Outcome Detection (HOD):} 
Given a sentence $s = w_1, \ldots,w_M$ extracted from a clinical trial abstract, the goal of HOD is to identify an outcome span $o_d = b_i, \ldots,b_N$ (i.e OSD), and subsequently predict a plausible outcome type $t(o_{d}) \in \cY$ for $o_d$ (i.e. OC), where $1 \leq i \leq N \leq M$, and $\cY$ is a predefined set of outcome types. 

We propose Label Context-aware Attention Model (LCAM), a sequence-to-sequence-to-set (\textsc{seq2seq2set}) model,  which uses a single encoder to represent an input sentence and two decoders --  one for predicting the label for each word in OSD and another for predicting the outcome type in OC. 
LCAM is designed to jointly learn contextualised label attention-based distributions at word- and sentence-levels in order to capture which label/s a word or a sentence is more semantically related to. 
We call them contextualised because they are enriched by global contextual representations of the abstracts to which the sentences belongs. 
Label attention incorporates label sparsity information and hence semantic correlation between documents and labels. 

A baseline BiLSTM and or clinically informed BERT$\mathrm{_{base}}$~\cite{devlin2018bert} models are used at the encoding stage of our model and later for decoding with sigmoid prediction layers. 
We also use a multi-label prediction (MLP) layer for the two tasks (i.e. OSD and OC), with a relaxed constraint at token-level that ensures only the top (most relevant) prediction is retained, whereas all predicted (relevant) outcome types are retained at the sentence-level during OC.
We use an MLP layer because some annotated outcomes belong to multiple outcome types.
 For example, \textit{depression} belongs to both \textit{``Physiological''} and \textit{``Life-Impact''} outcome types.

HOD remains a challenging task due to the lack of a consensus on how outcomes should be reported and classified~\cite{kahan2017comparison}.
\newcite{Dodd2018ADiscovery}~recently built a taxonomy to standardise outcome classifications in clinical records, which has been used to annotate the EBM-COMET~\cite{ebm-comet} dataset. 
Following these recent developments, we use EBM-COMET to align outcome annotations in the evaluation dataset we use in our experiments~\cite{Dodd2018ADiscovery}. 
Our main contributions in this work are summarised as follows\footnote{Our Code and datasets are located at \url{https://github.com/MichealAbaho/Label-Context-Aware-Attention-Model.git}}:
\begin{enumerate}
 \item We propose the Label Context-aware Attention Model to simultaneously learn label-attention weighted representations at word- and sentence-level. These representations are then evaluated on a biomedical text mining task that extracts and classifies health outcomes (HOD).
 \item We  introduce a flexible, re-usable unsupervised text alignment approach that extracts parallel annotations from comparable datasets. We use this alignment for data augmentation in a low-resource setting.
 \item We investigate the document-level contributions by a piece of text (e.g. an abstract) for predictions made at the token-level.
 \end{enumerate}

\section{Related work} \label{sec:relatedwork}

Joint training to achieve a dichotomy of tasks has previously been attempted, particularly for sequence labelling and sentence classification. 
Targeting Named Entity Recognition (NER) and Relation Extraction (RE), \newcite{chen2020joint} transfer BERT representations via a joint learning strategy to extract clinically relevant entities and their syntactic relationships. 
In their work, the joint learning models exhibit dramatic performance improvements over disjoint (standalone) models for the RE task. 
Our work differs from~\cite{chen2020joint} in that we use attention layers prior to the first and second classification layers.
\newcite{ma2017jointly} train a sparse attention-based LSTM to learn context features extracted from a convolution neural network (CNN). The resulting hidden representations are used for label prediction at each time step for sequence labelling, and subsequently aggregated via average pooling to obtain a representation for sentence classification. 
The sparse constraint is strategically biased during weights assignment (i.e. important words are assigned larger weights compared to less important words).

\newcite{karimi2020adversarial} perform ABSA~\cite{xu2019bert} by feeding a BERT architecture with a sentence 
$s = ([CLS],x_{1:j},[SEP], x_{j+1:n},[SEP])$, where $x_{1:j}$ is a sentence containing an aspect of a product, $x_{j+1:n}$ is a customer review sentence directed to the aspect and $[CLS]$ is a token not only indicating the beginning of a sequence, but also a sentiment polarity in the customer review about the aspect. 
They fine-tune a BERT model to conduct both aspect extraction and aspect sentiment classification.  
The above mentioned works tend to generate attention-based sentence-level representations that encapsulate the contribution each word would make in predicting sentence categories.
We however generate label-inclined attention representations at word-level that can be used to effectively deduce word categories/labels. To the best of our knowledge, we are the first to perform a joint learning task that achieves MLP at two classification stages, token- and sentence-levels, while using only the top predictions at token level.

\section{Data} \label{sec:data}
The absence of a standardised outcome classification systems prompted~\newcite{Nye2018ALiterature} to annotate outcomes with an arbitrary selection of outcome type labels aligned to Medical Subject Headings (MeSH) vocabulary.\footnote{\url{https://www.nlm.nih.gov/mesh}} 
Moreover their outcome annotations have been discovered with flaws in recent work~\cite{abaho2019correcting}, such as \emph{statistical metrics} and \emph{measurement tools} annotated as part of clinical outcomes e.g. \textit{``mean arterial blood pressure''} instead of \textit{``arterial blood pressure''}, \textit{``Quality of life Questionnaire''} instead of \textit{``Quality of life''}, \textit{``Work-related stress scores''} instead of \textit{``Work-related stress''}. 

Motivated by the taxonomy proposed by \newcite{Dodd2018ADiscovery} to standardise outcome classifications in electronic databases and inspired the annotation of EBM-COMET dataset~\cite{ebm-comet}, we attempt to align EBM-NLP's arbitrary outcome classifications to standard outcome classifications that  are proposed by~\newcite{Dodd2018ADiscovery}. These standard classifications were found (after extensive analysis and testing) to provide sufficient granularity and scope of trial outcomes. 
We propose an unsupervised label alignment method to identify and align parallel annotations across the EBM-NLP and EBM-COMET. Additionally, we use the discovered semantic similarity between the two datasets and merge them in order to create a larger dataset for evaluating our joint learning approach. The merged dataset contains labels that follow the taxonomy proposed by~\newcite{Dodd2018ADiscovery}. 
All three datasets are used during evaluation, with each one being randomly split into two, where 80\% is retained for training and 20\% for testing as shown in \autoref{tab:dataset_stats}. We hypothesise that the merged dataset would improve performance we obtain on the original independent datasets.

\begin{table}[]
\centering
\resizebox{\columnwidth}{!}{
\begin{tabular}{l|c|c|c}
\toprule
                         & EBM-COMET & EBM-NLP & \begin{tabular}[c]{@{}c@{}}EBM-COMET +\\ EBM-NLP\end{tabular} \\ \midrule
\# of Abstracts          & 300       & 5000    & 5300                                                          \\
\# of sentences          & 5193      & 40092   & 45285                                                         \\
\# of outcome labels     & 5         & 6       & 5                                                             \\
avg sentence length      & 21.0        & 26.0      & 25.0                                                            \\
\# of Training sentences & 4155      & 32074       & 36229                                                         \\
\# of Testing sentences  & 1038      & 8018       & 9056                                                         \\ \bottomrule
\end{tabular}
}
\caption{Datasets statistics rounded off to zero decimal}
\label{tab:dataset_stats}
\end{table}

\begin{table*}[h]
\centering
\resizebox{16cm}{!}{
\begin{tabular}{@{}l|c|c|ccccccccc|ccc|c@{}}
\toprule
                & Physiological                   & Mortality       & \multicolumn{9}{|c|}{Life-Impact}                                                                                                                                 & \multicolumn{3}{|c|}{Resource-use} & Adverse-effects \\ \midrule
                & P 0              & P 1             & P 25            & P 26            & P 27            & P 28            & P 29            & P 30            & P 31            & P 32            & P 33            & P 34         & P 35             & P 36               & P 38            \\ \midrule
Adverse-effects & 0.0615          & 0.1532          & 0.1226          & 0.1893          & 0.2001          & 0.1348          & 0.1169          & 0.2555          & 0.2320          & 0.0897          & 0.1936          & 0.2561       & 0.1768           & 0.1043& \textbf{0.0562} \\
Mental          & 0.0387          & 0.1829          & \textbf{0.0444} & \textbf{0.0928} & \textbf{0.1529} & \textbf{0.0623} & \textbf{0.0419} & \textbf{0.2214} & \textbf{0.1624} & \textbf{0.0624} & \textbf{0.1063} & \textbf{0.2537 } & 0.1955         & 0.1041       & 0.1904          \\
Mortality       & 01330           & \textbf{0.0187} & 0.1722          & 0.2562          & 0.2563          & 0.2171          & 0.1821          & 0.2594          & 0.2956          & 0.1559          & 0.2349          & 0.2855         & 0.1976           & 0.1905  & 0.2082          \\
Pain            & 0.0947          & 0.2310          & 0.1266          & 0.2181          & 0.1906          & 0.1316          & 0.1634          & 0.2662          & 0.2089          & 0.1290          & 0.2209          & 0.2770       & 0.2269          & 0.1422& 0.2096          \\
Physical        & \textbf{0.0114}  & 0.1582          & 0.0698          & 0.1494          & 0.1878          & 0.1126          & 0.0788          & 0.2363          & 0.2059          & 0.0639          & 0.1461          & 0.2539      & \textbf{0.1758}  & \textbf{0.0761}  & 0.1803          \\ \bottomrule
\end{tabular}
}
\caption{Cosine distance between representations of  EBM-NLP labels (first column) and EBM-COMET labels (top and second row).
EBM-COMET outcome type labels were drawn from the outcome domains defined in \cite{Dodd2018ADiscovery} taxonomy. 
Due to space limitations, we denote these domains as P X such as P 0, P 1 etc. The taxonomy hierarchically categorised them into 5 outcome types which are accordingly included in the top row. Outcome domains definitions are, P 0-Physiological/clinical, P 1-Mortality/survival, P 25-Physical functioning, P 26-Social functioning, P 27-Role functioning, P 28-Emotional
functioning/wellbeing, P 29-Cognitive functioning, P 30-Global quality of life, P 31-Perceived health status, P 32-Delivery of care, P 33-Personal circumstances, P 34-Economic, P 35-Hospital, P 36-Need for further
intervention, P 37-Societal/carer burden, P 38-Adverse events/effects}
\label{tab:inter-label-matrix}
\end{table*}

\subsection{Label alignment (LA) for Comparable Datasets} \label{sec:label-document-alignment}

Given two datasets $\cS$  and $\cT$ with comparable content, with $\cS$ containing $x$ labels such that $L_{s} = \{l_s^1, \ldots,l_s^x\}$ and $\cT$ containing $y$ labels $L_{t} = \{l_t^1,\ldots,l_t^y\}$, we design LA
to measure the similarity between each pair of labels $(l_s,l_t)$.

For this purpose, we first create an embedding for each label $l_s$ in a sentence $s(\in\cS)$ by applying mean pooling over the span of  embeddings (extracted using pre-trained BioBERT~\cite{lee2020biobert}) for the tokens corresponding to an outcome annotated with $l_s$ as shown in \eqref{eq:biobert}. Next, we average the embeddings of all outcome spans that are annotated with $l_s$ in all sentences in $\cS$ to generate an outcome type label embedding $\vec{l}_s$. Likewise, we create an outcome type label embedding, $\vec{l}_t$ for each outcome type in the target dataset $\cT$. After generating label embeddings for all outcome types in both $\cS$ and $\cT$, we compute the cosine similarity between each pair of $\vec{l}_s$ and $\vec{l}_t$ as the alignment score between each pair of labels $l_s$ and $l_t$ respectively.
\begin{align}
	\label{eq:biobert}
    \vec{O}_{l_s} = \frac{1}{d}\sum_{i}^{i+(d-1)} \mathrm{Biobert}(w_i)
\end{align}
where $O_{l_s}$, is an outcome span annotated with outcome type label $l_s$, $i$ and $i+(d-1)$ are the locations of the first and last words of the outcome span.
\begin{align}
    \vec{l}_s = \frac{1}{|l_s|}\sum_{1}^{|l_s|} \vec{O}_{l_s}
\end{align}
where $|l_s|$ is the number of outcome spans annotated with label $l_s$ and $\vec{l}_s$ is label $l_s$ embedding.

\autoref{tab:inter-label-matrix} shows the similarity scores for label pairs $(l_s,l_t)$ across $\cS$ (EBM-COMET) and $\cT$ (EBM-NLP) respectively. 
For each label (which is an  outcome domain) in EBM-COMET, we identify the EBM-NLP label which is most similar to it by searching for the least cosine distance across the entire column.
After identifying those pairs that are most similar, we automatically replace outcome type labels in EBM-NLP with EBM-COMET outcome type labels as informed by the similarity measure.

Results show that Physiological outcomes (containing domain P 0) are similar to Physical outcomes and therefore the latter outcomes are labelled Physiological, Life-Impact outcomes are similar to Mental outcomes and therefore the latter outcomes are labelled Life-Impact. Mortality and Adverse-effects outcomes both remain unchanged because both categories exists in source and target datasets, and their respective outcomes are discovered to be similar. We evaluate the LCAM architecture on the resulting merged dataset, and additionally, evaluate the alignment approach by comparing the performances before and after merging. 

\section{Label Context-aware Attention Model}
\begin{figure*}[t]
\centering
\includegraphics[width=\textwidth]{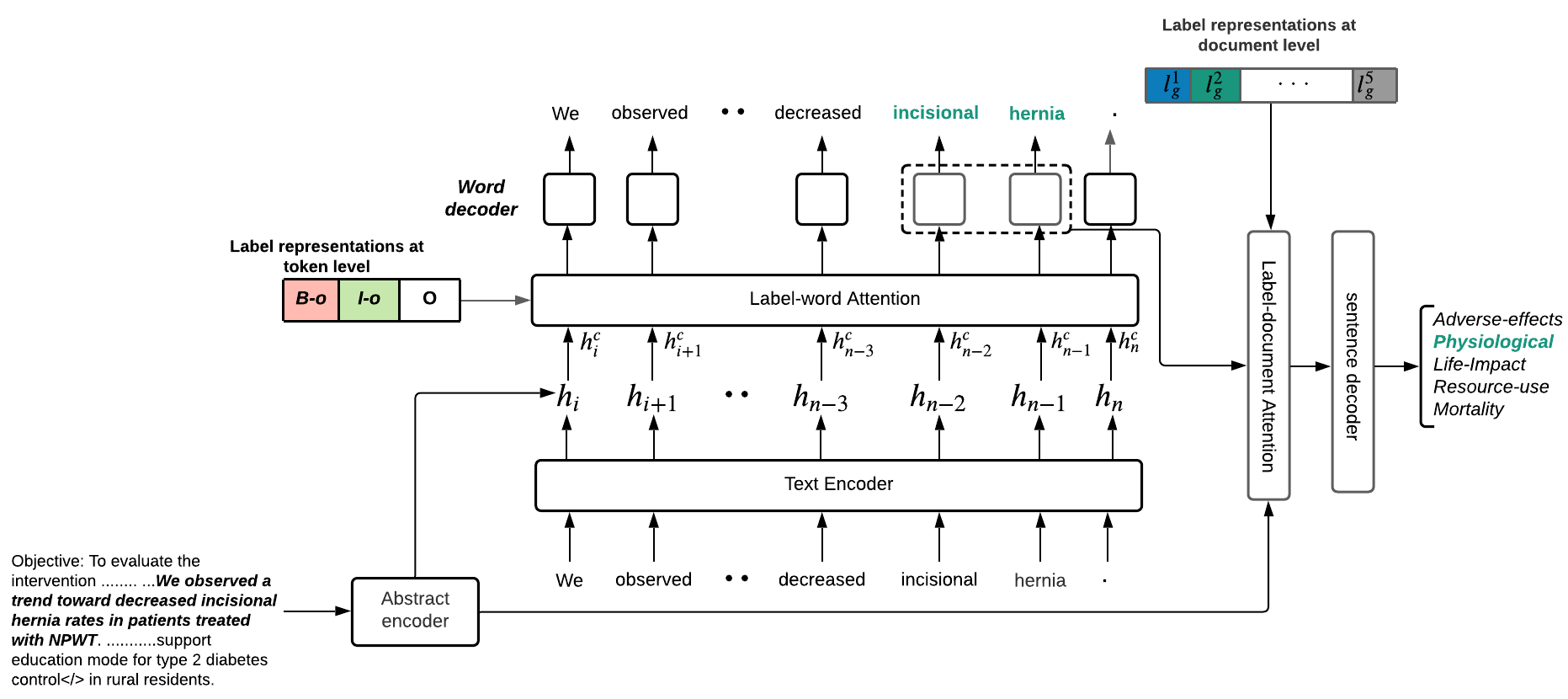}
\caption{Illustration of the LCAM Architecture. It encodes a sequence of tokens of a sentence within an abstract, generates contextualised representations by adding a global representation of the abstract at word- and sentence-level. Two attention layers are used to aid generation of label-aware representations used to decode labels at word-level for OSD and sentence-level for OC.}
\label{fig:lcam}
\end{figure*}

\autoref{fig:lcam} illustrates an end-to-end \textsc{seq2seq2set} architecture of the LCAM model. 
It depicts a two-phased process to achieve classification at token and sentence level. In phase 1, input tokens are encoded into representations which are sent to a decoder (i.e. a sigmoid layer) to predict a label for each word, hence OSD. Subsequently, in phase 2, the token-level representations are used to generate individual outcome span representations, which are sent to another decoder (sigmoid layer) that is used to predict the label/s for each outcome span, hence OC. 
We use MLP for the OC task because some outcomes are annotated with  multiple outcome types.
The pseudo code for LCAM is shown in the Supplementary.

\subsection{Outcome Span Detection (OSD)}
\label{sec:pdf}

Given a set of sentences $\cS = \{s_i\}_{i=1}^{|\cS|}$ within an abstract $a$, each $s_i$ having $N$ words, $s_i = w_1,\ldots,w_N$, with each word tagged to a label $l_w$ and use BIO tagging scheme~\cite{sang1999representing}.
OSD aims to extract one or more outcome spans within $s_i$. For example, in  \autoref{fig:lcam}, OSD extracts the outcome span \textit{``incisional hernia''} given the input sentence.

\paragraph{Encoder:}
In our OSD task setting, we initially implement a baseline LCAM using a BiLSTM to encode input tokens (that are represented by $d$-dimensional word embeddings we obtain using GloVe~\cite{pennington2014glove}\footnote{\url{https://github.com/stanfordnlp/GloVe}}) into hidden representations for every word within an input sentence. We then consider generating each input words hidden representation using a pre-trained clinically informed BERT$\mathrm{_{base}}$ model called BioBERT~\cite{lee2020biobert}. The LCAM model learns \eqref{eqn:eqlabel},
\begin{align}
\label{eqn:eqlabel}
\begin{split}
 \vec{h}_n = \mathrm{BiLSTM}(w_n) ,
\\
 \vec{h}_n = \mathrm{BioBERT}(w_n) 
\end{split}
\end{align}
where $w_n \in s_i$, $\vec{h}_n \in \R^{k \times 1}$ and $k$ is the dimensionality of the hidden state. The upper equation under \ref{eqn:eqlabel} is used for a BiLSTM Text encoder and the lower for a BioBERT one.

\subsection{Abstract Hidden State Context}
\label{sec:layout}


To make the hidden state representation context-aware, we add a compound representation of the abstract in which the sentence containing $w_n$ belongs.
\begin{align}
    \vec{h}^{c}_{n} = \vec{h}_n + f(\mathrm{AbsEncoder}(a))
\end{align}
where $f$ is a function computing the average pooled representation of the encoded abstract, $\mathrm{AbsEncoder} \in \{\mathrm{BiLSTM, BioBERT}\}$,  $\mathrm{AbsEncoder}(a) \in \R^{k \times |a|}$, $|a|$ is the length of the abstract (measured by the number of tokens contained in it) and $f(\mathrm{AbsEncoder}(a)) \in \R^{k \times 1}$.

\subsection{Label-word attention}
We compute two different attention scores, the first is to enable the model pay appropriate attention to each word when generating the overall outcome span representation. Then the second attention score, is to allow the words interact with the labels in order to capture the semantic relation between them, hence making the representations more label-aware.  
To obtain the first attention vector $\vec{A}^{(1)}$, we use a self-attention mechanism~\cite{al2018hierarchical,lin2017structured} that uses two weight parameters and a hyper parameter $b$ that can be set arbitrary,
\begin{align}
\label{eq:attention1}
    \vec{A^{(1)}_n} =  \mathrm{softmax}(\mat{W}\tanh(\mat{V}\vec{h}_n^c))
\end{align}
where $\mat{W} \in \R^{|l_w| \times b}$, $\mat{V} \in \R^{b \times k}$ and $\vec{A^{(1)}} \in \R^{|l_w| \times 1}$. $|l_w|$ is the number of token-level labels.
Furthermore, we obtain a label-word attention vector $\vec{A}^{(2)}$ using a trainable matrix $\mat{U}\in \R^{|l_w| \times k}$. Similar to the interaction function \newcite{du2019explicit} use, this attention is computed in \eqref{eq:attention2} as the dot product between the $\vec{h_n^c}$ and $\mat{U}$,

\begin{align}
    \label{eq:attention2}
    \vec{A}^{(2)}_n = \mat{U}\vec{h}_n^c
\end{align}
where $\vec{A}^{(2)}_n \in \R^{|l_w| \times 1}$. 

\paragraph{Label-word representation}
The overall representation used by the decoder for classification of each token is obtained by merging the two attention distributions from the previous paragraphs as shown by \eqref{eq:labe-word-rep},
\begin{align}
    \label{eq:labe-word-rep}
    \vec{E}_n^{t_l} = \vec{A}_n^{(1)}\vec{h}_n^{c^\top} + \vec{A}_n^{(2)}\vec{h}_n^{c^\top}
\end{align}
where $\vec{E}_n^{t_l} \in \R^{|l_w| \times k}$, denotes the token-level ($t_l$) representation.
The training objective is to maximise the probability of a singular ground truth label and minimise a cross-entropy loss,
\begin{align} \label{eq:gps_to_cartesian}
L_{osd} = -\sum_{n=1}^{N}\sum_{i=1}^{|l_w|}y_{n,i}\log(\hat{y}_{n,i}).
\end{align}
where 
$N$ is number of tokens in a sentence, $l_w$ is the number of labels.

\subsection{Outcome Classification (OC)}
OC predicts outcome types for the outcome spans extracted during OSD. 
Similar to what is done at token-level, 
we add an abstract representation (which is a mean pool of its token's representations) 
to add context to each tokens representation. 
An outcome span is represented by concatenating the vectors of its constituent words, 
\begin{align}
\vec{O}_s = \bigoplus_{i=1}^m(\vec{E}_i^{t_l}+f(\mathrm{AbsEncoder}(a)))
\end{align}

where $m$ is the number of tokens contained in outcome span $O_s$.
We adopt the aforementioned self-attention and label-word attention methods at sentence-level to aid extraction of an attention based sentence-level representation of an outcome as follows:
\begin{align}
    \vec{E}_s^{s_l} = \vec{A}^{(1)}\vec{O}_s + \vec{A}^{(2)}\vec{O}_s
\end{align}
where $[\vec{A}^{(1)}, \vec{A}^{(2)}] \in \R^{|l_s| \times m}$, $\vec{O}_s\in\R^{m \times k}$ and $s \ge 0$. 
Given an outcome span representation $\vec{E}^{s_l}$, the training objective at sentence-level ($s_l$) is to maximize the probability of the set of terms,
\begin{align}
  \underset{\theta}{\argmax}P(y=(l_s^1,l_s^2,...,l_s^6) \in l_s|\vec{E}^{s_l};\theta)
\end{align}
{\small
\begin{align}
    L_{oc} \,=\, -\sum_{i=1}^{|l_s|}y_i\log(\hat{y_i}) + (1-y_i)\log(1-\hat{y_i})
\end{align}
}
where $y_i \in \{0, 1\}$, $\hat{y_i} \in [0, 1]$ $l_s \in$ \{Physiological, Mortality, Life-Impact, Resource-use, Adverse-effects\}. The overall joint model loss is:
\begin{align}
    L = L_{osd} + L_{oc}
\end{align}

\section{Experiments}

\begin{table*}[!htb]
\centering
\resizebox{15cm}{!}{
\begin{tabular}{@{}lllllllll@{}}
\toprule
                  & Task          &            & \multicolumn{3}{c}{OSD}                                                         & \multicolumn{3}{c}{OC}                                                         \\ \midrule
Dataset           & Model         & setup      & \multicolumn{1}{c}{P}    & \multicolumn{1}{c}{R}    & \multicolumn{1}{c}{F}    & \multicolumn{1}{c}{P}    & \multicolumn{1}{c}{R}    & \multicolumn{1}{c}{F}    \\ \midrule \midrule
EBM-COMET         & Baseline      & Joint      & \multicolumn{1}{c}{63.0} & \multicolumn{1}{c}{55.0} & \multicolumn{1}{c}{59.0} & \multicolumn{1}{c}{78.0} & \multicolumn{1}{c}{73.0} & \multicolumn{1}{c}{74.0} \\
& BioBERT       & Standalone & 74.0                         & 74.3                         & \textbf{74.2}                &   76.7                       &78.4                          & 77.5                       \\
                  & SCIBERT       & Standalone & 72.3                         & 72.9                         &  72.6                &   76.3                       &78.1                          & 77.2                       \\
                  & LCAM-BioBERT & Joint      & 73.0                     & 64.0                     & 68.0                     & 83.0                     & 76.0                     & \textbf{83.0}            \\
                  
EBM-NLP           & Baseline      & Joint      & 49.0                     & 40.0                     & 44.0                     & 65.0                     & 59.0                     & 61.0                     \\
& BioBERT       & Standalone & 48.2                      & 51.5                        & 49.8                       &    65.7                      &  74.6                        &       \textbf{69.9}                   \\
                  & SCIBERT       & Standalone & 48.5                         & 49.7                         & 49.1                &   64.2                       &66.5                          & 65.3                       \\
                  & LCAM-BioBERT & Joint      & 57.0                     & 49.0                     &\textbf{51.0}                     & 67.0                     & 65.0                     & 66.0           \\
                  
EBM-COMET+EBM-NLP & Baseline      & Joint      & 62.0                     & 54.0                     & 58.0                     & 68.0                     & 64.0                     & 65.0                     \\
 & BioBERT       & Standalone &    58.6                      &  61.4                        & 60.0                         &    81.4                      &     83.0                     &  \textbf{82.2}    \\
                  & SCIBERT       & Standalone & 56.2                        & 62.3                         & 59.1                &   73.4                       &75.7                          & 74.5                      
                  \\ 
                  & LCAM-BioBERT & Joint      & 61.0                     & 61.0                 &\textbf{61.0}                     & 78.0                     & 72.0                     & 75.0                     \\
                 \bottomrule
\end{tabular}
}
\caption{Outcome span detection (OSD) and Outcome classification (OC) results in terms of F1 on the three datasets. Baseline, is a LCAM architecture with a BiLSTM sequence encoder.}
\label{tab:main-results}
\end{table*}

The joint learning LCAM framework is evaluated on the three datasets discussed in \autoref{sec:data}: the expertly annotated EBM-COMET, the EBM-NLP~\cite{Nye2018ALiterature} and the merged dataset created by aligning (covered in section \autoref{sec:data}) parallel annotations between EBM-NLP and EBM-COMET.

\subsection{Implementation}
For pre-processing the data, we first label each word in the sentences contained in an abstract with either one of $\{B,I,O\}$.
Subsequently, to the end of each sentence, we include a list of outcome types corresponding to the outcome spans in the sentence. However, it is important to note that, not all sentences within an abstract had outcome spans.
For example, the annotated sentence below contains outcome span ``Incisional hernia'' whose outcome label (Physiological) is placed at the end of the sentence. 

\textit{
``We/\textbf{[O]} observed/\textbf{[O]} a/\textbf{[O]} trend/\textbf{[O]} toward/\textbf{[O]} decreased/\textbf{[O]} incisional/\textbf{[B-outcome]} hernia/\textbf{[I-outcome]} rates/\textbf{[O]} in/\textbf{[O]} patients/\textbf{[O]} treated/\textbf{[O]} with/\textbf{[O]} NPWT/\textbf{[O]} ./\textbf{[O]}''. [[\textbf{Physiological}]] }

We tuned hyper-parameters using 20\% of the training data of the merged dataset (EBM-NLP+EBM-COMET) as a development set. The optimal settings included,
a batchsize of 64, dropout of 0.1, 10 epochs, hidden state dimension for the BiLSTM and BioBERT encoders was set to 300 and 768 respectively. For the BioBERT model, we used features from BioBERT's ultimate layer, a practice that has been endorsed in the past~\cite{naseem2020biomedical, yoon2019pre, hao2020investigating}. 
We use the Adam optimizer~\cite{kingma2014adam} with a learning rate of 0.001.
Experiments were performed using a Titan RTX 24GB GPU. 

\begin{table*}[!htb]
\centering
\resizebox{13cm}{!}{
\begin{tabular}{@{}lcccccc@{}}
\toprule
          &      & OSD   &      &      & OC   &      \\ \midrule
          LCAM-BioBERT & P    & R    & F    & P    & R    & F    \\ \midrule \midrule
EBM-COMET & 73.0/83.0 & 64.0/64.0 & 68.0/71.0 & 83.0/90.0 & 76.0/80.0 & 83.0/84.0 \\
EBM-NLP   & 57.0/60.0 & 49.0/47.0 & 51.0/53.0 & 65.0/76.0   & 65.0/72.0 & 64.0/74.0 \\ \bottomrule
\end{tabular}
}
\caption{Effect of dataset merging via label alignment. For each dataset, we report the performance on its test split obtained by LCAM-BioBERT trained on the corresponding train split (shown on the left side of /) vs. 
on the merger of the train splits of EBM-COMET and  EBM-NLP (shown on the right side of /).}
\label{tab:alignment}
\end{table*}

\subsection{Setup}
The Joint setup is concurrent sequence labelling (OSD) and sequence classification (OC) whereas the standalone setup, is OSD and OC performed separately. The former is achieved using (a) a Baseline model, LCAM-BiLSTM (using a BiLSTM encoder) (b) LCAM-BioBERT (using  BioBERT encoder), whereas the latter is achieved by fine-tuning the original (c) BioBERT and (d) SciBERT~\cite{beltagy2019scibert} models.
Our datasets are novel in the sense that the outcome type labels of the outcomes are drawn from~\newcite{Dodd2018ADiscovery} taxonomy, which is not the basis of prior outcome annotations such as the EBM-NLP dataset. 
The models were evaluated on the tasks by reporting the macro-averaged F1. For the standalone models, we use token-classification and text-classification fine-tuning scripts provided by Huggingface~\cite{wolf-etal-2020-transformers} for OSD and OC respectively.
Inaddition to the macro-F1, we visualise ranking metrics pertaining to MLP, in order to compare our model to related work for MLP. The metrics of focus include precision at top n P@n (fraction of the top n predictions that is present in the ground truth) and Normalized Discounted Cumulated Gain at top n (nDCG@n). 
\subsection{Results}
\vspace{-1mm}
The first set of results we report in \autoref{tab:main-results} are based on the independent test sets (\autoref{tab:dataset_stats}) for each of the datasets. The joint LCAM-BioBERT and standalone BioBERT models are not only competitive but they consistently outperform the baseline model for both OSD and OC tasks. 
We observe the LCAM-BioBERT model outperform the other models in the OSD experiments for the last two datasets in \autoref{tab:main-results}.
On the other hand, the standalone BioBERT model achieves higher F1 scores for the last two datasets in the OC task. 

\begin{figure*}[!htb]
\centering
\includegraphics[width=11cm, height=7cm]{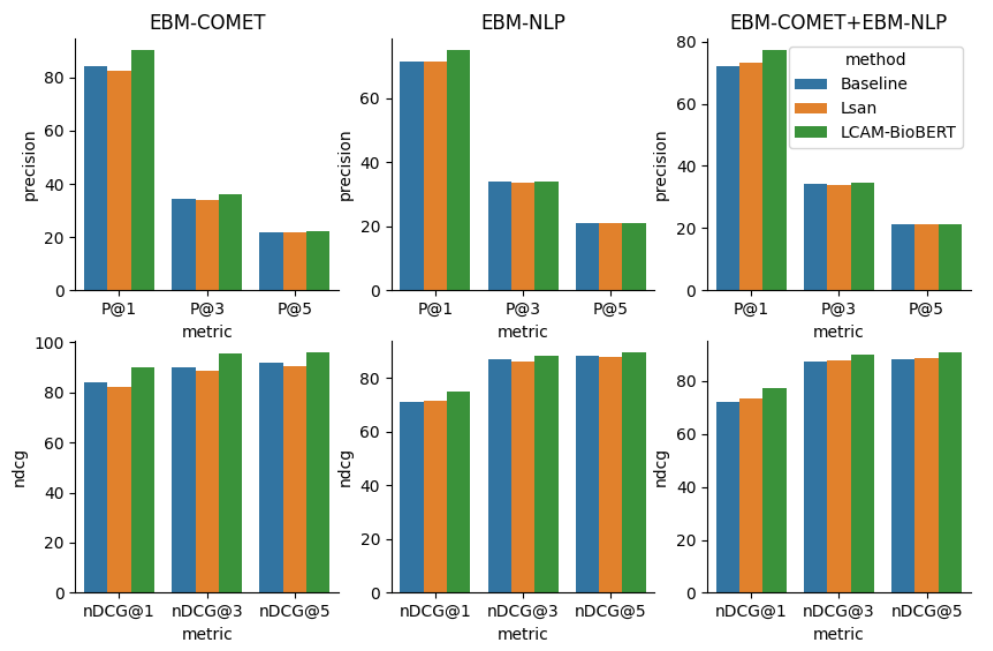}
\vspace{-2mm}
\caption{P@n and nDCG@n for three datasets}
\label{fig:ranking}
\end{figure*}

\subsubsection{Impact of the abstract context injection and Label attention}
\begin{table}[t]
\centering
\resizebox{\columnwidth}{!}{
\begin{tabular}{@{}lrcc@{}}
\toprule
          & LCAM                  & OSD(F)   & OC(F)   \\ \midrule \midrule
EBM-COMET &  - Attention                     & -10.0& -12.0   \\
          & - Abstract                            & -3.0 & -5.0 \\
EBM-NLP   &  - Attention & -9.0 & -7.0 \\
          & - Abstract                            & -7.0 & -2.0 \\
EBM-COMET &  - Attention & -11.0& -15.0    \\
+EBM-NLP  & - Abstract                            & -3.0 & -1.0 \\ \bottomrule
\end{tabular}
}
\caption{OSD and OC performance percentage decline  when either the attention mechanism or the abstract representation are eliminated from the joint learning model (LCAM-BioBERT).}
\label{tab:at-ab}
\end{table}

As shown in \autoref{tab:at-ab}, the performance deteriorates (with respect to the results reported in \autoref{tab:main-results}) without the attention layers (``- Attention'')  by averagely 10\% 
for OSD and
11.3\% for OC.
Similarly, exclusion of the abstract representation (``- Abstract'') leads to an average performance decline of 4.3\% for OSD and 2.7\% for OC. As observed the decline resulting from ``- Abstract'' is less significant than that resulting from ``- Attention'' for both OSD and OC tasks. 


This decline explains the significant impact of both (1) the semantic relational information between both tokens and labels as well as outcome spans and labels gathered by the attention mechanism, (2) information from the text surrounding a token or an outcome span embedded into an abstract representation. This therefore justifies inclusion of both these components. 

To evaluate the proposed label alignment method (\autoref{sec:label-document-alignment}), we train a model using the aligned dataset (EBM-COMET+EBM-NLP) and evaluate it on the test sets of the original datasets in \autoref{tab:alignment}.
We see significant improvements in F-scores for OSD in both EBM-COMET and EBM-NLP.
Additionally, for OC, we see a significant improvement in F-score on EBM-NLP dataset and a slight improvement in F-score on the EBM-COMET dataset. Overall, this result shows that the proposed label alignment method enables us to improve performance for both OSD and OC tasks.

\begin{table*}
\centering
\resizebox{15cm}{!}{
\begin{tabular}{@{}llll@{}}
\toprule
& Example Input sentence          & Predicted labels  & Predicted labels        \\                                                            
        & & \multicolumn{1}{c}{P@1}  & \multicolumn{1}{c}{P@2}  \\
                                                     \midrule
Ground truth                                           & \begin{tabular}[c]{@{}l@{}}The primary outcomes were {\textcolor{BlueGreen}{\textbf{hospitalised death}}}$^1$,  \textcolor{Maroon}{\textbf{severe disability}}$^2$ at 15 months of age, \\ \textcolor{Maroon}{\textbf{neonatal behavioural neurological}}$^3$ assessment (nbna) score at 28 days of age, and Bayley \\ scales of \textcolor{Maroon}{\textbf{infant development}}$^4$ (BSID) score (including \textcolor{Maroon}{\textbf{mental development}}$^5$ index (mdi) \\ score and \textcolor{Maroon}{\textbf{psychomotor development}}$^6$ index (pdi) score) at 15 months of age at follow-up. \end{tabular} & \begin{tabular}[c]{@{}l@{}}1. Mortality\\ 2. Life-Impact\\ 3. Life-Impact\\ 4. Life-Impact\\ 5. Life-Impact\\ 6. Life-Impact\end{tabular} & \\ \\

\begin{tabular}[c]{@{}l@{}}LCAM\\ Output\end{tabular} & \begin{tabular}[c]{@{}l@{}}The primary outcomes were hospitalised \textcolor{BlueGreen}{\textbf{death}}$^1$,  \textcolor{OliveGreen}{\textbf{severe}}$^2$ \textcolor{Maroon}{\textbf{disability}}$^3$ at 15 months of age, \\ neonatal behavioural neurological assessment (nbna) score at 28 days of age, and Bayley \\ scales of infant development (BSID) score (including mental \textcolor{Maroon}{\textbf{development}}$^4$ index (mdi) \\ score and \textcolor{Maroon}{\textbf{psychomotor development}}$^5$ index (pdi) score) at 15 months of age at follow-up. \end{tabular} & \begin{tabular}[c]{@{}l@{}}1. Mortality\\ 2. Physiological\\ 3. Life-Impact\\ 4. Life-Impact\\ 5. Life-Impact\end{tabular}  & \\ 

\midrule
Ground truth                                           & \begin{tabular}[c]{@{}l@{}}These results confirm retrospective studies and add that histopathology subtype is a strong \\ determinant of  \textcolor{Plum}{\textbf{disease-free survival (DFS)}}$^1$, in resected MAGE-A3-positive MSCLC.  \end{tabular} &  \begin{tabular}[c]{@{}l@{}}1. Physiological \end{tabular} & \begin{tabular}[c]{@{}l@{}}1. Mortality \end{tabular} \\  \\

\begin{tabular}[c]{@{}l@{}}LCAM\\ Output\end{tabular} & \begin{tabular}[c]{@{}l@{}}These results confirm retrospective studies and add that histopathology subtype is a strong \\ determinant of  \textcolor{Plum}{\textbf{disease-free survival}}$^1$ (DFS), in resected MAGE-A3-positive MSCLC. \end{tabular} & \begin{tabular}[c]{@{}l@{}}1. Physiological \end{tabular}  & \begin{tabular}[c]{@{}l@{}}1. Mortality \end{tabular} \\ 

\midrule
Ground truth                                           & \begin{tabular}[c]{@{}l@{}}The duration of total \textcolor{Cyan}{\textbf{hospital stay}}$^1$, and  \textcolor{Cyan}{\textbf{postoperative hospital stay}}$^2$ in the ag \\ (10.86 +/- 5.64, 5.69 +/- 4.55) d were significantly shorter than that in the cg (.10.86 +/- 5.64, \\ 5.09 +/- 4.55) d (p=0.01, p=0.01)) \end{tabular} & \begin{tabular}[c]{@{}l@{}}1. Resource-use\\ 2. Resource-use \end{tabular} & \\ \\

\begin{tabular}[c]{@{}l@{}}LCAM\\ Output\end{tabular} & \begin{tabular}[c]{@{}l@{}}The duration of total \textcolor{Cyan}{\textbf{hospital}}$^1$ \textcolor{OliveGreen}{\textbf{stay}}$^2$, and \textcolor{OliveGreen}{\textbf{postoperative}}$^3$ \textcolor{Cyan}{\textbf{hospital stay}}$^4$ in the ag \\ (10.86 +/- 5.64, 5.69 +/- 4.55) d were significantly shorter than that in the cg (.10.86 +/- 5.64, \\ 5.09 +/- 4.55) d (p=0.01, p=0.01)) \end{tabular} & \begin{tabular}[c]{@{}l@{}}1. Resource-use\\ 2. Physiological\\ 3. Physiological\\ 4. Resource-use\end{tabular}  & \\ 

\bottomrule
\end{tabular}
}
\caption{Sample error predictions made by the joint learning model, with coloured words representing the outcome phrase (both in ground truth and output) and the colours representing different outcome types which are output. For multi-label predictions, we include P@1 and P@2 to indicate the top most predictions for the outcome phrase in question such as in example 2.}
\label{tab:error-analysis}
\end{table*}


To further evaluate the LCAM-BioBERT model, we focus on the OC task results alone where the classifier returns the outcome types given an outcome span, and
compare MLP performance to the baseline and another related MLP model, label-specific attention network (LSAN)~\cite{xiao2019label}, that learns biLSTM representations for multi-label classification of sentences. For comparison, we compute P@n and nDCG@n using formulas similar to~\cite{xiao2019label}. 
As illustrated in \autoref{fig:ranking}, the LCAM model outperforms its counterparts for all datasets, and most notably for P@1. 
Our joint BiLSTM baseline model performs comparably with LSAN, and indeed outperforms it on the EBM-COMET dataset for P@1, nDCG@1 and nDCG@3. 
We attribute LCAMs superior performance to (1) Using a domain-specific (biomedical) language representation model (BioBERT) at its encoding layer, (2) Applying label-specific attention prior to classifying a token as well as before classifying the mean pooled representation of an outcome span and finally (3) injecting global contextual knowledge from the abstract into the token and document (outcome-span) representations.

\subsubsection{Error Analysis}

We review a few sample instances that exhibit the mistakes the joint LCAM model makes in the OSD and OC tasks in \autoref{tab:error-analysis}. 

\paragraph{OSD errors:} We observe the model partially detecting outcome phrases e.g. 
In Example 1, it detects death instead of hospitalised death, development instead of mental development, and in Example 2, it does not detect ``(DFS)'' as apart of the outcome phrase. Additionally, it completely misses some outcomes such as infant development in Example 1.

\paragraph{OC errors:} Incorrect token-level predictions will most likely result into incorrect outcome classification. 
In Example 1, Instead of severe disability, the model detects ``severe'' as an outcome and ``disability'' as a separate outcome and classifies them as Physiological and Life-Impact respectively. 
Similarly, in Example 3, both outcomes are misclassified because at token level multiple outcomes are detected rather than one, hospital and stay rather than hospital stay, postoperative and hospital stay rather than postoperative hospital stay.

\section{Conclusion}


We proposed a method to jointly detect outcome spans and types using a label attention approach. Moreover, we proposed a method to align multiple comparable datasets to train a reliable outcome classifier. Given real-world scenarios where it is often impractical or computationally demanding to build a model for each and every single task, our experimental results demonstrate the effectiveness of an approach that simultaneously (jointly) achieves two different task without compromising the performance of the individual tasks when decoupled.

\section{Ethical Considerations}
Joint learning can have multiple applications, where multiple tasks are simultaneously achieved whilst preserving (or even improving) standalone performance when tasks are separately conducted. In this particular work, we are motivated by the need to jointly model a pair of tasks (Outcome span detection and Outcome classification) in order to enhance outcome information retrieval. Recent developments in the domain such as emergence of an outcome classification system that is aimed at standardising outcome reporting and classification motivated us to re-construct the datasets we use in order to align them with this classification. The datasets contain text from abstracts of clinical trials published on PubMed. We cannot ascertain  that all these abstracts are unbiased assessments of effects of interventions, especially with recurring articles citing several biases including \textit{selection bias} (trial clinicians favour certain participating patients because of personal reasons), \textit{reporting/publishing bias} (only reporting statistically significant results) and many more. Nevertheless, we provide more details and reference these datasets both within the article and the supplementary material. 

\bibliographystyle{acl_natbib}
\bibliography{ms}

\section*{Appendices}
\appendix

\title{Learning to jointly label and classify sequences using label-specific attention in a low resource domain \\ -- Supplementary Materials --}



\date{}

\maketitle
\appendix
\section{Joint learning using LCAM}
To demonstrate the flow of our joint learning training, we use the pseudo code in algorithm \ref{alg:algorithm1} to show how we arrive at the joint model loss. For each token's hidden state (line 8), we compute a context aware hidden state by adding to it an encoded abstract representation line 9 and then compute two attention scores (line 10 - 14) that both capture the contribution the token makes to each label. These are then used to generate a label-word representation (line 16), all label-word representations forming a sentence (line 17) are used to compute an outcome extraction(OE) loss using eqn 9 (line 19). Once again we add context to the newly generated toke-level representations (line 20). For every outcome, we repeat steps in lines 10-14 to obtain label attention scores., i.e. depicting the contribution the particular outcome phrase makes to each label and these are used to obtain a label-document representation for the outcome (line 30). This representation is then used to compute the outcome classification loss (line 32). The loss we minimise in the joint learning is computed as shown by line 33.

\begin{algorithm}[!htb]
\caption{LCAM Training}
\label{alg:algorithm1}
\begin{algorithmic}[1] 
\STATE \textbf{Input}: train data, \textbf{Output}: model weights
\FOR{abstract $a$ in train data}
    \STATE Obtain $Abs=\mathrm{AbsEncoder}(a)$ 
    \FOR{sent $s$ in $a$}
        \STATE Obtain $\mat{H}=Encoder(s)$
        \STATE where $\mat{H} \in \R^{k\times n}$ 
        \STATE \textbf{Initialise:} an empty tensor $\mathrm{S}$
        \FOR{$\vec{h}_n$ in $\mat{H}$}
            \STATE $\vec{h}_n^c = \vec{h}_n + f(Abs)$ 
            \STATE {\small Obtain $\vec{A}^{(1)}=\mathrm{softmax}(\mat{W}\tanh{\mat{V}\vec{h}_n^c})$} 
            \STATE where $\vec{V}\in \R^{b \times k}$, $\mat{W} \in \R^{|l_w| \times b}$,
            \STATE and $\vec{A}^{(1)} \in \R^{|l_w| \times 1}$
            \STATE Obtain $\vec{A}^{(2)} = \mat{U}\vec{h}_n^c$ 
            \STATE where $\mat{U} \in \R^{|l_w| \times k}$, $\vec{A} \in \R^{|l_w|\times1}$ 
            \STATE \textbf{label-word representation}: 
            \STATE $\vec{E}^{t_l} = \vec{A}^{(1)}\vec{h}_n^{c^\top} + \vec{A}^{(2)}\vec{h}_n^{c^\top}$ 
            \STATE  $\mathrm{S} = \mathrm{S} \oplus \vec{E}^{t_l}$ 
        \ENDFOR
        \STATE Compute Loss eqn 9 - $L_{osd}$
        \STATE $\forall \vec{E}^{t_l} \in \mathrm{S}: \vec{E}^{t_l} = \vec{E}^{t_l} + f(Abs)$
        \STATE  $\forall O_{x} \in \mathrm{S}$, where $x \ge 0$ \& $O_{x} \in \R^{m \times k}$ 
        \STATE i.e. outcome $\vec{O}_x$ has $m$ tokens
        \FOR{outcome $\vec{O}$ in $\mathrm{S}$}
            \STATE Obtain {\small $A^{(1)}=\mathrm{softmax}(\mat{W}\tanh(\vec{V}\vec{O}^\top))$} 
            \STATE where $\vec{V} \in \R^{b \times k}$, $\mat{W} \in \R^{|l_s| \times b}$
            \STATE and $\vec{A} \in \R^{|l_s| \times m}$
            \STATE Obtain $\vec{A}^{(2)} = \mat{U}\vec{O}^\top$ 
            \STATE where $\mat{U} \in \R^{|l_s| \times k}$, $\vec{A} \in \R^{|l_s| \times m}$ 
            \STATE \textbf{label-document representation of an outcome}: 
            \STATE $\vec{E}^{s_l} = \vec{A}^{(1)}\vec{O} + \vec{A}^{(2)}\vec{O}$ 
        \ENDFOR
        \STATE Compute Loss $L_{oc}$ eqn 13 
        \STATE  minimise  model loss $L = L_{osd} +  L_{oc}$
    \ENDFOR
 \ENDFOR
\end{algorithmic}
\end{algorithm}

\section{Hyperparameters and Run time}
We perform a grid search through multiple combinations of hyperparameters included in \autoref{tab:parameter_joint} below. Using 20\% of EBM-COMET+EBM-NLP dataset as a dev set, we obtain the best F1 values.  \autoref{tab:parameter_joint} shows the range of values (including the lower and upper bound) for which the LCAM-BioBert is tuned to obtain optimal configurations. Using a shared TITAN RTX 24GB GPU, the baseline joint model i.e. LCAM-BiLSTM runs for approximately 45 minutes when evaluating on the EBM-COMET dataset, 190 minutes when evaluating on the EBM-NLP dataset and at-least 320 minutes on the merged dataset EBM-COMET+EBM-NLP. For the LCAM-BioBERT model, the experiments last at-least 14 hours on the EBM-COMET dataset, 30 hours on the EBM-NLP and 42 hours on the merged EBM-COMET+EBM-NLP. 

\autoref{tab:parameter_stand} includes the tuned ranges for the Standalone models (BioBERT and SciBERT) which we fine-tune for the outcome extraction (OE) and outcome classification task. Similar to the joint model, the best values are chosen based on the EBM-COMET test set F1 values. Training and evaluation on the EBM-COMET, EBM-NLP and EBM-COMET+EBM-NLP consume 7, 34, and 45 GPU hours respectively.
\begin{table}[!htb]
\centering
\resizebox{\columnwidth}{!}{
\begin{tabular}{@{}lcc@{}}
\toprule
\textbf{Parameter}                                                             & \textbf{Tuned-range}                                                                & \textbf{Optimal}                                   \\ \midrule
Batch size                                                                     & {[}16,32,64{]}                                                                      & 64                                                 \\
Drop out                                                                       & {[}0.1,0.2,0.3,0.4,0.5{]}                                                           & 0.1                                                \\
\begin{tabular}[c]{@{}l@{}}Embedding dim\\ \hspace{3.5mm}-Baseline\\ \hspace{3.5mm}-BERT models\end{tabular} & \begin{tabular}[c]{@{}c@{}}\_\\ \_\end{tabular}                                     & \begin{tabular}[c]{@{}c@{}}300\\  768\end{tabular} \\
b                                                                              & {[}150, 200, 250{]}                                                                 &                                                    \\
Optimizer                                                                      & {[}Adam, SGD{]}                                                                     & Adam                                               \\
Epochs                                                                     & {[}5,10,15{]}                                                                     & 10  \\
Learning rate                                                                  & \begin{tabular}[c]{@{}c@{}}{[}5e-4, 1e-4, 5e-3, 1e-3, \\ 5e-2, 1e-2{]}\end{tabular} & 1e-3                                               \\ \bottomrule
\end{tabular}
}
\caption{Parameter settings for the joint models}
\label{tab:parameter_joint}
\end{table}

\begin{table}[!htb]
\centering
\resizebox{\columnwidth}{!}{
\begin{tabular}{@{}lcc@{}}
\toprule
\textbf{Parameter}                                                             & \textbf{Tuned-range}                                                                & \textbf{Optimal}                                   \\ \midrule
Train Batch size                                                                     & {[}8,16,32{]}                                                                      & 16,32                                                 \\
Eval Batch size                                                                     & {[}8,16,32{]}                                                                      & 8                                                \\
Embedding dim                                                                       & \_                                                           & 768                                                \\
Optimizer                                                                      & {[}Adam, SGD{]}                                                                     & Adam                                               \\
Epochs                                                                     & {[}5,10,15{]}                                                                     & 10  \\
Learning rate                                                                  & \begin{tabular}[c]{@{}c@{}}{[}5e-5, 1e-4, 5e-3, 1e-3{]}\end{tabular} & 5e-5                                               \\ \bottomrule
\end{tabular}
}
\caption{Parameter settings for the Standalone models}
\label{tab:parameter_stand}
\end{table}

\section{Datasets}
\subsection{EBM-NLP}
EBM-NLP corpus~\cite{Nye2018ALiterature} is a crowd sourced dataset in which ca.5,000 clinical trial abstracts were annotated with elements in the health literature searching PICO framework~\cite{Huang2006EvaluationQuestions}. PICO stands for Participants, Interventions, Comparators and Outcomes. The dataset has supported clinicalNLP research tasks~\cite{beltagy2019scibert, brockmeier2019improving}. The corpus has two versions, (1) the ``\textbf{starting spans}'' in which text spans are annotated with the literal ``PIO'' labels (I and C merged into I) and (2) the ``\textbf{hierarchical labels}'' in which the annotated outcome ``PIO'' spans were annotated with more specific labels aligned to the concepts codified by the Medical Subject Headings (MeSH)~\footnote{\url{https://www.nlm.nih.gov/mesh}}, for instance the Outcomes (O) spans are annotated with more granular (specific) labels which  
include Physical, Pain, Mental, Mortality and Adverse
effects. For the clinical recognition task we attempt, we use the hierarchical version of the dataset. The dataset has however been discovered to have flawed outcome annotations~\cite{abaho2019correcting} such as (1) statistical metrics and measurement tools annotated as part of clinical outcomes e.g.``\textit{mean arterial blood pressure}'' instead of ``\textit{arterial blood-pressure}'',``\textit{Quality of life Questionnaire}'' instead of ``\textit{Quality of life}'' and (2) Multiple outcomes annotated as a single outcome ``\textit{Systolic and Diastolic blood-
pressure}'' instead of ``\textit{Systolic blood-pressure}'' and ``\textit{Diastolic blood-pressure}''.

\subsection{EBM-COMET}
A biomedical corpus containing 300 PubMed ``Randomised controlled Trial'' abstracts manually annotated with outcome
classifications drawn from the taxonomy proposed by~\cite{Dodd2018ADiscovery}. The abstracts were annotated by two experts with extensive experience in
annotating outcomes in systematic reviews of clinical
trials~\cite{ebm-comet}. \newcite{Dodd2018ADiscovery}'s taxonomy hierarchically categorised 38 outcome domains into 5 outcome core areas and applied this classification system to 299 published core outcome sets (COS) in the Core Outcomes Measures in Effectiveness (COMET) database.

\subsection{EBM-COMET+EBM-NLP} 
We merge the two datasets above for two main purposes, (1) to align the annotations of the EBM-NLP to a standard classification system~\cite{Dodd2018ADiscovery} for outcomes and (2) create a larger dataset to use in evaluating our joint learning approach.

\subsection{Pre-processing}
We create one single vocabulary using the merged dataset and use it for all three datasets. Whilst generating the vocabulary, we simultaneously split the abstracts into sentences using Stanford tokeniser. This vocabulary is then used in creating tensors representing sentences, where each tensor contains id's of the token/words in the sentence. The same procedure is followed to create tensors containing id's of the labels (``BIO'') corresponding to the words in the sentences. Additionally, we create tensors with id's of outcome classification labels, so for each sentence tensor, their is a corresponding token-level label tensor and a sentence-level label (outcome label) tensor. For the baseline where we use a BiLSTM to learn GloVe representations, we follow \href{https://github.com/stanfordnlp/GloVe}{instructions} to extract GloVe\footnote{\url{https://github.com/stanfordnlp/GloVe}} specific vectors for words, token-level labels and sentence labels in the dataset. All the files with d-dimensional vectors are stored as .npy files. For the joint BERT-based models, we use flair~\cite{akbik2019flair} to extract TransformerWord Embeddings from pre-trained BioBERT for the tokens.




\end{document}